\documentclass[10pt,final]{article}

\usepackage{snlp2007}
\usepackage{times}
\usepackage{epsfig}
\usepackage{booktabs}
\usepackage{dialogue} \usepackage{abbrevs}
\usepackage[T1]{url}
\let\em\itswitch

\newif\ifpdf
\ifx\pdfoutput\undefined \else
  \ifx\pdfoutput\relax
  \else
    \ifcase\pdfoutput
    \else
      \pdftrue
    \fi
  \fi
\fi

\newabbrev\dt{Drill Tutor}

\summary{
Speaking a language and achieving \emph{proficiency} in another one is a
highly complex process which requires the acquisition of various kinds
of knowledge and skills, like the learning of words, rules and patterns
and their connection to communicative goals (intentions), the usual
starting point. To help the learner to acquire these skills we propose
an enhanced, electronic version of an age old method: \emph{pattern
drills} (henceforth PDs). While being highly regarded in the fifties,
PDs have become unpopular since then, partially because of their lack of
grounding (natural context) and rigidity. Despite these shortcomings we
do believe in the virtues of this approach, at least with regard to the
acquisition of basic linguistic reflexes or skills (automatisms),
necessary to survive in the new language. Of course, the method needs
improvement, and we will show here how this can be achieved. Unlike
tapes or books, computers are open media, allowing for dynamic changes,
taking users' performances and preferences into account. Building an
electronic version of PDs amounts to building an open resource,
accomodatable to the users' ever changing needs.
}

\begin{document}

\title{Let's get the student into the driver's seat}
\author{
\begin{tabular}[t]{c}
\large{Michael Zock}\hspace*{1cm}\large{Stergos D. Afantenos}\\
\\
LIF, CNRS (UMR 6166)\\
Universit\'e de la M\'editerran\'ee, Facult\'e des Sciences de Luminy \\
163, Avenue de Luminy - Case 901, 13288 Marseille C\'edex 9 - France \\
\{\texttt{michael.zock,stergos.afantenos}\}\texttt{@lif.univ-mrs.fr}
\end{tabular}
}

\maketitle

\section{Introduction}\label{sec:introduction}
Spontaneous speech is a cyclic process involving a loosely ordered set
of tasks: conceptual preparation, formulation, articulation. Given a goal one
has to decide what to say (conceptualization) and how to say it
(formulation), making sure that the chosen elements, words, can be
integrated into a coherent whole (sentence frame) and conform to the
grammar rules of the language (syntax, morphology). During vocal delivery (articulation), in itself already a quite demanding task, the speaker may
decide to initiate the next cycle, namely starting to plan the subsequent
ideational fragment.

Obviously, smooth execution of such a complex task requires not only
access to a huge library of ready made fragments in more or less abstract or
concrete form (patterns vs. words, or larger units), but also excellent organizational skills. Speed and knowledge are not all; proficient speakers are also flexible, capable to change on the fly incompatible elements
(dynamic accommodation). Not everything is necessarily planned in
advance, local adjustments may become necessary.

If speaking is already a complex task, to do so in a foreign language can be
even more daunting or overwhelming. There are at least three, probably
related reasons for this: lack of knowledge, lack of assurance and lack of
remembrance. Indeed, learning to speak a new language requires not only
learning a stock of new words and rules, but also to have the necessary confidence to
dare to speak, which supposes, of course, quick access
(for example, words) and remembrance of what has been learned.

To achieve these goals (increase/consolidation of knowledge,
fixation/memorisation, boosting of confidence) we have enhanced an
age-old method, pattern drills, by building an
electronic version of it. While the drill tutor (henceforth DT) is built
for learning Japanese, we believe that the method is general enough to
be applied to other languages.

PDs are a special kind of exercise based on notions like: analogy, task
decomposition (small steps), systematicity, repetition and feedback. Important as they may be, PDs, or exercices in general, are but
one of the many tools teachers rely on for teaching a language.
Dictionaries, grammars, video and textbook being other
means. None of them, except the first one  will be taken into consideration
in this paper. PDs are typically used in audio-oral lessons. Such lessons are generally composed of the following steps: 1) Presentation of a little drama, involving people trying to solve a communication problem at a given
place and time (hotel, train station). The student hears the story and is encouraged to play
one of the characters; 2) Presentation of contrastive examples for rule induction. 3) A phase of rule fixation. This is where the PDs come into play. 4) Re-use of the learned rules or pattern in a new
and similar, yet different situation. These four stages fulfill,
roughly speaking, the following functions (a) symbol grounding, i.e. illustration of the pragmatic usage of the structure;
(b) conceptualization, i.e. explanation/understanding of the rule (c)
memorization/automation of the patterns, and (d) transposition/consolidation of the learned material.

Obviously, there are many ways to learn a language, yet, one of them has proven to be quite efficient, at least for survival
purposes: PDs.\footnote{After having been very popular for many years, PDs
and instrumental conditioning on which they rely upon have been
discredited by linguists (see Chomsky's violent criticism
\cite{Chomsky} of Skinner's book \textit{Verbal Behavior}, an attempt
to provide an operational account of language), and more directly by
psychologists and pedagogues
\cite{Rivers64,Savignon}. While we do agree with
these criticisms, when the process of language production or the
architecture of the human mind are at stake, we do not share them at
all, when habit formation or the acquisition of automatisms are the
learning goal. For this specific task we do believe that principled
ways of staging the repetition of stimulus-response patterns together
with feedback are a valuable learning method. Interestingly enough,
patterns have been rehabilitated by one of Chomsky's best known 
students \cite{Jackendoff93}.}

Since PDs are neither a new nor an uncontroversial method, let us show
how some, if not practically all, of their shortcomings can be overcome.

Linguists describe languages in terms
of rules, but people hardly ever learn such descriptions, at least not
at the initial stages of acquiring a new language. What people do
learn though are patterns complying with these rules.\footnote{In our case, a pattern can be seen as the frozen instance of a step in the derivational process. Which step we want to focus on depends on the task (describing data, support the language user). The main function patterns serve in this context is to support the speaker at the next step/level in the process, whatever this level may be. Hence, patterns can be produced by a generative grammar, and they may be hybrid.}
This is definitely the case for beginners and this holds both for first and
second language acquisition. R. Weir \cite{Weir.62} provided evidence for this by showing that children do spontaneously what we do when being asked, rehearse linguistic structures.  Recording her daughter at  bedtime, she heard her doing spontaneously, what we do in school: drilling patterns. This kind of behavior has been confirmed by other studies. As the learner makes progress, i.e.
acquires more knowledge about the language, s/he will see the
limitations of the pattern (overgeneralisation), possibly refining it such as to
accommodate for the exceptions. Of course, people learn not only
patterns, but also the situations (context) in which they occur. The
latter can be seen as goals: seeing someone \emph{introduce himself}, hearing
him \emph{ask for a favor} or \emph{offering help}, the learner realizes that
the person s/he is observing uses over and over the same pattern though not necessarily always in the same situation. Given this tight connection between means
(patterns) and ends (goals), we have decided to integrate it in our
 DT: the fact that patterns are indexed in
terms of goals, allows the user to choose the means (patterns) as a
function of the end (goal, input). People are generally little motivated
to do something, unless they perceive its use, that is, the end a given action is serving for (means).

As pointed out elsewhere \cite{Zock&Quint.04b}, one of the drawbacks of
traditional PDs is their material support. Books or audio
cassettes being closed media, everything they contain has to be
anticipated in all its details prior to their release. Yet anticipation
can never be perfect for at least two reasons. Different people have
different needs (for example, the specific words someone would like to
learn, i.e. with which to drill a given pattern), and peoples' needs
change over time. This being so, we must take 
individual differences and the users' ever evolving needs into account. Yet
this is precisely what is precluded in the case of closed media. They do
not accept any change, update or accommodation after the product's
release. In addition they ignore user preferences, or personal learning
history: the road to ``success'' (the method offered) is the same for all,
forcing the learner into a straightjacket which, technologically
speaking, is not justified anymore. Indeed, none of these constraints
are very problematic for open media like computers. The lexical values
with which to instantiate the patterns' variables can be changed at any
moment, so can the order of examples, the speed of their presentation and the
number of repetitions be changed at will. As a
result, the same exercise can be used by a much larger group, or over a
 longer period of time.

\begin{figure*}[t]
\begin{center}
  \ifpdf
    \includegraphics[width=0.75\textwidth]{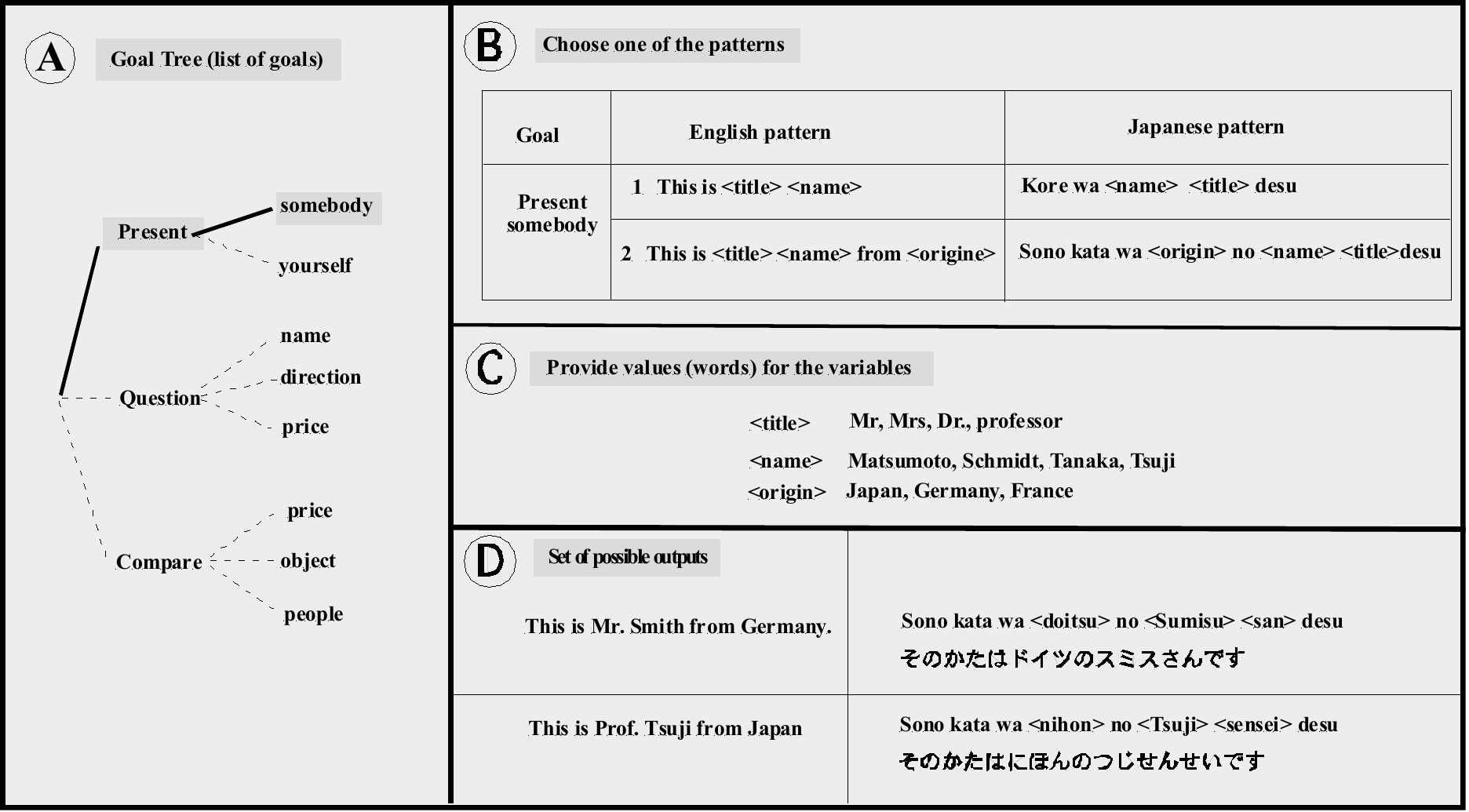}
  \else
    \includegraphics[width=0.75\linewidth]{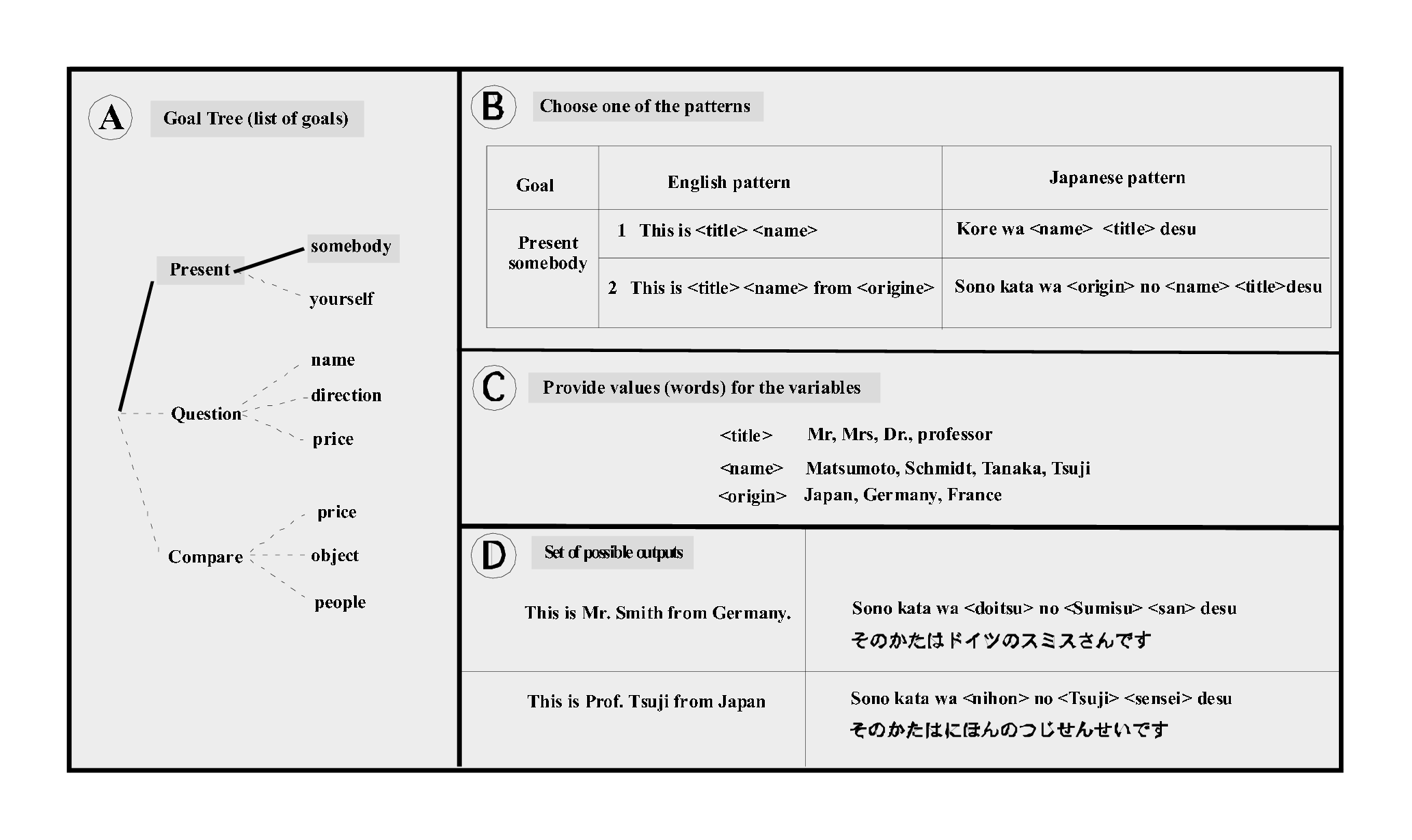}
  \fi
  \caption{The basic interaction process between the DT and the student.}
  \label{fig:walkthrough}
\end{center}
\end{figure*}

\section{An example of the Process}\label{sec:interaction}
Before showing how the resource is built, let us see how it is meant to work and in what respect it differs from conventional PDs used in a language lab. Let's start with the latter, illustrating it with the simplest case, substitution drills requiring no morphological changes.

The student receives a model, which could be composed of a question (let's say, ``what is this?''), a stimulus (``a pen'') and the answer (``This is a pen''). From then on, he will only be given the stimulus and feedback concerning his answer (the machine producing the correct sentence). Of course, the user has to produce the answer in the first place.

While being similar to classical PDs, our approach is nevertheless quit a bit different in various respects. First of all, it is the student who chooses the pattern he'd like to work on, as he knows (arguably) best what his needs are. Next, we have indexed patterns in terms of goals. This is necessary in order to allow the learner to find the pattern he'd like to work on. With the system growing, grows the list of patterns. Hence, access becomes an issue. Also, associating patterns with goals allows the student to realize the pattern's communicative function. Third, since we don't have a parser or speech recognizer, we have a problem concerning the user's output. Actually, the learner does not type at all. Since the focus is on speed (fluency), we'd like to avoid slowinging down the process by having the input provided via the keyboard. Hence we ask the user to produce the sentence mentally or to speak it out loud, and to check then whether their result corresponds to the system's output. Fourth, the system's output is also written. This can be considered as a disadvantage, yet it can also be turned into a big advantage if, as planned, a speech synthesizer is added. Doing so would allow not only to discover grapheme-phoneme mappings, hence support the learning of reading and writing, but also allow to support memorization by showing intonation curves and to control the speed of the vocal output, which leads us to the last point. Unlike tapes, which are a closed medium, we can change at any moment certain parameters like (a) speed; (b) order of examples, (c) number of repetitions after which an element is taken from the list (staging of repetitions and maximal presentation of problematic cases), etc.

Figure~\ref{fig:walkthrough} here above illustrates the way how the student gets from the starting point, a goal (frame A), to its linguistic realization, the endpoint (frame D) by using the DT. The process is initiated via the choice of a goal (\emph{present somebody}, step 1, frame A) to which the system answers with a set of patterns (step 2, frame B). The user chooses one of them (B1 vs. B2, step 3), signalling then the specific lexical values with which he would like the pattern to be instantiated (frame C, step 4). The system has now all the information needed to create the exercise (frame D), presenting sequentially a model,\footnote{The latter is basically composed of a sentence (step 1), a stimulus (the lexical value of the variable, step 2), and the new sentence based on the model and the stimulus (step 3, see Table 1).} the stimulus (chosen word), followed by the student's answer and the system's confirmation/information (normally also a sentence, implying that the student's answer is correct if the two sentences match and incorrect in the opposite case). The process continues until the student decides to stop, or until s/he has done all the exercices.

\begin{table}[htb]
  \centering\scriptsize
  \begin{tabular}{l|l}
    \toprule
    \multicolumn{2}{c}{\emph{Goal : Present Somebody}} \\
    \midrule
    Pattern in English    & \footnotesize\verb|This is <title>| \\
                          & \footnotesize\verb|<name> from <origin>.| \\
    Pattern in Japanese   & \footnotesize\verb|Kono kata wa <origin>| \\
                          & \footnotesize\verb|no <name> <title> desu.| \\
    \midrule
    Stimulus              & \footnotesize\verb|Mr, Schmidt, Germany| \\
    Instantiated Sentence & \footnotesize\texttt{Kono kata wa \underline{doitsu}} \\
                          & \footnotesize\texttt{\underline{no Shimito} \underline{san} desu.} \\
    \bottomrule
  \end{tabular}
  \caption{Model given by the system}
  \label{table:system_model}
\end{table}

\noindent Note that, if the values of the variables $<$title$>$, $<$name$>$, $<$origin$>$ were (professor, Tsuji, Japan) rather than (Mr, Smith, Germany), then the outputs would vary, of course, accordingly (see Table~\ref{table:interaction}).

\begin{table}[htb]
  \centering\scriptsize
  \begin{tabular}{l|l}
    \toprule
    \multicolumn{2}{c}{\emph{Goal : Present Somebody}} \\
    \midrule
    \textsc{System} (stimulus) & \footnotesize\verb|Prof, Tsuji, Japan| \\
    \textsc{User} (response)   & \footnotesize\texttt{Kono kata wa \underline{nihon} no} \\
                          & \footnotesize\texttt{\underline{Tsuji} \underline{sensei} desu.} \\
    \textsc{System} (verification) & correct \\
    \bottomrule
  \end{tabular}
  \caption{Interaction with the user in the training session}
  \label{table:interaction}
\end{table}

Two points might be worth mentioning: (1) Conceptual input is distributed over time, specification taking place in several steps: first by choosing the abstract overall structure or pattern (steps 1 and 2) and then by providing the variables' concrete lexical values: 'Mr, Mrs, Dr, or professor' for the variable $<$title$>$ (step 4). (2) At the moment, we do not rely on any morphological component which obviously limits the system's scope. Any question-answer pair or sentence transformation may require such a component, and we will surely integrate it later on. For the time being, the idea is just to illustrate the system's basic mechanism and the interaction between the system and the student.


\section{Current State of the Drill Tutor}
The DT is a client/server web application written in
PHP\footnote{\url{http://www.php.net/}} and sitting atop an Apache
server\footnote{\url{http://httpd.apache.org/}} in a Linux machine.
The DT is divided into two separate areas: one reserved for data
acquisition linking words, patterns and goals (the ``Expert Area''),
the other being reserved for exercising (the ``Student Area''). The
implemented operations are as follows.

\subsection{Expert Area}\label{sec:impl:expert}
\paragraph{\footnotesize{User Authentication}}
People entitled to make changes to the database (henceforth called experts)
have to authenticate themselves, via a user name and a password. This
is necessary in order to make sure that people work on their own data
and to avoid  inconsistencies/noise in the database.

\paragraph{\footnotesize{Creation of New Patterns and Goals}}
Currently the system has about 30 patterns.\footnote{Actually, the
number of patterns is not really what counts at this stage, as the focus
is on the implementation of an editor designed for building,
modifying and using a database. Also, while it would have been easy to
copy patterns from one of the many text books, we have refrained from
doing so, not only for reasons of copyright, but also for reasons of
metalanguage. The terms in which these patterns are defined are neither always consistent nor very felicitous.} To allow for quick access, patterns
need to be indexed. This is done here via goals. Of course, other
criteria could or should be used. In order to provide new data
(typically a new goal and its associated patterns and words) the
expert can either use the system's graphical user interface (henceforth GUI), or
upload a file containing the necessary information. This implies in
the first case \emph{(a)} naming the goal and specifying its parent
(see also the next paragraph),
\emph{(b)} providing, in the source and target languages, all patterns
likely to achieve this  goal, and \emph{(c)} providing values (words)
for the variables. Once this is done, the system presents the expert
all possible sentences computable on the basis of the input, allowing
him to check them for well formedness. After this the expert can provide
additional input. If the data are communicated via a file, care must
be taken that the latter complies with the syntactic rules.

\paragraph{\footnotesize{Structuring Goals into Trees}}
Learners knowing their needs prefer to make their own choices rather than being told what patterns to work on. To do so, we must give them
the means to express their needs, or, in this particular case, to locate the patterns. We do this via an index (goals). Indexing can be done from various points of view (pragmatic,
semantic, syntactic). We have chosen the first (pragmatic), leaving
syntactic/semantic indexing (i.e. composing the message) for later on.
As goals can contain other goals, we have a tree or a
hierarchy of goals. Hence, in order to find the wanted
pattern, the student has to navigate in such goal tree (see also
Figure~\ref{fig:walkthrough}). To enable the system to create this kind of
structure, experts have to state with their input where in the
hierarchy fits their new goal and its associated pattern(s). This is
done via the parent node.
\paragraph{\footnotesize{Modification of Goals}}
The data given to the system (goals, patterns and lexical values)
can be modified at any time. In other words, experts can add, delete or
modify the patterns and values for any goal  inserted.
\paragraph{\footnotesize{Visualization of the database}}
The data given can be visualized as a table which can be useful for
checking completeness and consistency of the patterns, or for appreciating the adequacy of
the metalanguage.
\paragraph{\footnotesize{Creating Backups}}
Computers are notorious for crashing, causing users to loose
the work of days or even months. To circumvent this we allow experts to create backups of
their work, so that it can be restored at any point in time.
\paragraph{\footnotesize{Providing a new interface language}}
The learning of a new
language should  be independent of the used interface language, i.e. the language in which the information relevant for teaching/learning takes place.
In order to accomplish this, the DT allows the expert to translate the interface and  the linguistic data into a chosen source
language (usually, the learner's mother tongue). The same holds true for the transliteration table, whose
equivalents of the Hiragana symbols have to be given in the
new language. This can easily be done via the GUI.

\subsection{The Student Area}\label{sec:impl:student}
\paragraph{\footnotesize{Working on the exercises}}
As mentioned already, the students can choose the
pattern they'd like to work on. To find the wanted pattern they navigate
in the goal hierarchy.
Once they've found the goal they will be presented with all its associated patterns, meaning that they have to choose again, though,
this time from a much smaller set. Of course, they could also choose to
work with all of the goals, but this is rather atypical at the
beginning stages. Regardless of the user's choice (one vs. several patterns), the process develops as follows. First, students are shown an
example sentence (model) in which a single element will be replaced. In
the simplest case (substitution drill), only one element will be
replaced, no morphological changes taking place. Next are given, in
random order, the elements (stimulus) to be inserted into the proper slot. Doing so should help the students not only to memorize words, but also use (or produce) them in the proper syntactic context.

After this, students are shown the
correct sentence both in roman characters as well as in their
transliterated Japanese form (for the time being only in Hiragana). The
process iterates until the student has acquired the patterns or has
decided to stop.
It should be noted that the transliteration of the Hiragana
appears in the interface language chosen by the user. If it were Greek,
then the
transliterated sentence would also be in Greek characters, next to the
Hiragana, of course.

The whole process is presented in a sober graphical user
interface. In addition, users are allowed to define keyboard shortcuts,
using keystrokes rather than moving the mouse over radio buttons. This increases speed and confort for telling the system that one
has been able to produce the expected output, information
necessary to decide whether a given combination (pattern + specific
word) should be kept on the exercise list.

\paragraph{\footnotesize{Monitoring of Errors}}
Of course, the whole process would be of little use if the students
were not given some means (feedback) to assess the quality of their work.
Actually, the system keeps track of the users' performance (errors made
during the training session),
presenting  them at the end  statistics concerning the patterns they
have worked on. This allows the students to devote more time to
problematic cases.
\paragraph{\footnotesize{Exercises for specific problems}}
One of the many problems foreigners have when learning
Japanese is their counting system, as the words expressing numerical values depend not only on these values, but also on the nature (or certain features) of the counted object. To this end, quanfiers (-bon, -mai, -biki, etc.) are added to the counters. While western languages use the same word, let's say 'three' to talk about given number of 'pencils, tickets, or dogs', japanese use a different quantifier for each case : san-bon, san-mai, san-biki, etc. In order to account for this fact, we have created exercises allowing for
selective variation of  the number, the object or both. Similarly, we
have created specific exercises for learning the expression of time or family
relationships.
\paragraph{\footnotesize{Let the students use their own
metalanguage}}
At some point experts will have to decide how to refer to a goal or a variable, that is, how to call or name them. Any of the following could be used to refer to a given object: \emph{subject}, \emph{noun }, \emph{food }. Still, students might not like any of them.
After all, what counts is that they can find what they are looking
for, and to this end they have to rely on (or navigate by using)
concepts that are meaningful to them. Hence, we should give them the
means to name things the way they want, possibly even using several
terms. In order to allow for this we provide a GUI allowing students to
change the names of the variables and goals to
their likings. With respect to the implementation, this
information is kept in form of cookies on the learner's side.
\paragraph{\footnotesize{Choosing the interface language}}
As mentioned already in section~\ref{sec:impl:expert}, experts  have
the possibility to provide other interface languages than the default one,
English. This allows people to study Japanese via a
 language they feel most comfortable with.
If it were Russian, all the menus, scripts (transliterated
Hiragana), goals and patterns would also be in this language at the
GUI level.

\section{Future work}\label{sec:future}
There are several ways to speed up data acquisition and conceptual input: (a) integration of a multilingual dictionary, (b) automatic detection of pattern instances in corpora and (c) automatic pattern abstraction on the basis of concrete input (sentences).

\subsubsection*{Integration of a Dictionary}
With every new goal the expert has to provide not only its associated patterns, but also a list of values for the variables' values along with their translation. This puts a lot of work on the experts' shoulders. To alleviate this burden we could integrate an electronic dictionary. A multilingual resource like Papillon (\url{http://www.papillon-dictionary.org/Home.po}) with its dozen of languages would definitely be a good candidate. Adding an electronic dictionary to the DT is but the first step towards the creation of a richer environment, enabling the expert to provide example pattern(s) for a given goal, and to find good lexical candidates to fill the patterns' variables. The chosen words would, of course, be automatically translated  into the ``target language''.

\subsubsection*{Searching for Instances of Patterns in a corpus}
When providing a pattern for a goal, the expert has also to provide  a list of \emph{values} with which to fill the pattern's variables. Imagine the following pattern, along with its equivalent in Japanese (see Table~\ref{table:patterns_web}).

\begin{table}[htb]
  \centering\scriptsize
  \begin{tabular}{ll}
    \toprule
    \textbf{English Pattern} & \textbf{Japanese Pattern} \\
    \midrule
    A : \ttfamily  What is that? & A : \ttfamily soreha nan desu ka\\
    B : \ttfamily This is a $<$object$>$. & B : \ttfamily koreha $<$object$>$ desu\\
    \bottomrule
  \end{tabular}
  \caption{A pattern in English and Japanese.}\label{table:patterns_web}
\end{table}

The following could be a list of candidate values : desk (tsukue), chair (isu), lamp (denki), etc. Even if this list is open, the real problems are errors and scaling. With the list growing, grows the danger of making errors. Anyhow, keying-in words is certainly not a very enticing task. This being so, we have decided to offer the expert another solution: instead of having him imagine and type in all these words, we provide him with a set of candidates from which he can choose. The method is quite simple. Knowing the goal, we know the pattern, and knowing the pattern we can use it, filtering a corpus to find instances of the variables. In other words, searching potential values of the variables of a pattern amounts to making a \emph{generic grep-like} search over a well chosen text (say, the electronic version of a text book, a teaching method, etc.), replacing the variables by the values' \emph{wild-card} characters \texttt{*}. Thus, the search for values for the pattern of Table~\ref{table:patterns_web} could take the following form: ``\texttt{This is a *}''.

Obviously, the success of this approach hinges critically on the quality of the corpus and (the generality of) the query. Indeed, if the query is too general, we will get too many hits, which, even though compatible with the pattern, they are not really what the expert is looking for. For example, the pattern here above could yield not only ``this is a \textit{house}'', which is fine, but also  ``this is a \textit{QBitmap object}'', and  ``this is a \textit{chair that Louis XIV used to have in one of his bedrooms}'', which are by no means what we were looking for.

The procedure just described could be useful not only for the expert, but also for the student. Indeed, the same techniques could be used to search for instances of a pattern, i.e. a ``similar'' sentences \emph{in context} to the one provided as input. To this end the system could perform a grep search in a corpus, large enough to yield interesting examples, while being sufficiently accessible to be meaningful for the user. Hence, rather than resorting to the web which contains too much noise, we could use as corpus texts that the user is familiar with, composed of sentences he has either encountered or is likely to understand. The coursebook used for teaching the language would be a good candidate. This would not only expand the students' experience of the patterns and the language, by illustrating words in the context of sentences and discourse, it would also help them to memorize the words by learning new connections.

\subsubsection*{Guessing Patterns from concrete Sentences}
Here we would like to take the whole process one step further and describe how users could get the system to guess the pattern they have in mind when producing a specific utterance. The input could even be given in the users' mother tongue, provided that the lexical database knows the corresponding lexical values. Suppose you produced "quiero una cerveza", wanting the system to understand, not only literally that you'd like a beer, but more generically, that  ``there is a \textit{person}, who \textit{desires} to get a certain kind of \textit{drink}'', in order to retrieve then, as discussed already, instances of this pattern. 

There are at least three issues here at stake: (1) determine the elements to be replaced by a variable (in theory, nearly all words could be replaced); (2) determine the adequate level of naming the variable (while the notion of NP is meaningful for linguists, categories like ``people, fruit, means of transportation,'' etc. are far more meaningful for ``ordinary'' people.); (3) massage the data (by annotating them) to ensure that the program will find instances of the abstracted word (category, variable).

\subsubsection*{Enrichment of the Database}
At the moment of writing this paper, we have mainly been concerned with the implementation of the basic infrastructure. Though little effort has been put until now into actually feeding the database with data (goals, patterns and values), this is vital information. While there are many sources we can draw upon \cite{Chino,Kamiya}, we must make careful choices. Increasing the number of patterns is not all, we must also make sure to choose only those that are really useful. In addition, we should be careful about the metalanguage which ideally should be meaningful even for the linguistically innocent users.

\section{Conclusions}\label{sec:conclusions}
Becoming fluent in a language requires not only learning words and methods for accessing them quickly, but also learning how to place them and how to make the necessary morphological adjustments. This is not a small feat, considering that all this has to be done fast, and on top of it, content must be planned.

The work presented here is the result of less than 12 months' work. It is implemented in PHP, and the supported languages are English at the interface level and  Japanese as the language to be learned. We plan to add other languages, a speech synthesizer and a morphological component. We also intend to make the system available on the web in the near future.

Having linked patterns to goals should help users to perceive the function of a given structure (i.e which goal(s) can be reached by using a particular pattern). Yet, most importantly, this linkage offers the possibility to get instances of the pattern from a document (corpus). This is interesting not only for data acquisition (building the resource by feeding it with lexical entries likely to occur in a given pattern), but also for remembrance. In addition, presenting patterns with new material allows expanding the learner's experience of the language. The fact that most goals are associated with multiple patterns allows to extend the range of the exercise, reducing thus boredom. Instead of drilling one single pattern in response to a chosen goal, the system can prompt the user by presenting him various patterns.

Obviously, PDs are not a panacea, yet used in the right way they can do wonders. Just like a tennis player might want to go back to the court and train his basic strokes, a language learner may feel the need to drill resisting patterns. We must beware though that patterns are just one element of a long chain. They need to be learned, but once interiorized they must be placed back into the context where they have come from, a real communicative scene. Without this additional experience they will simply fail to produce the wanted effect, that is, help us achieve our communicative goals.

Computers are a medium escaping many of the constraints (rigidity, closedness) other media (tapes or books) are condemned to. They allow for variable order of presentation, dynamic updating of words and much more. Learning a language does not mean memorizing sentences, actually, we tend to forget those sooner or later. What usually remains are ideas, words and patterns, rather than full fledged sentences and rules. Hence, forgetting sentences is not a problem anymore, since we know now how to build them. This is the goal to which the DT contributes.

\bibliographystyle{snlp2007}
\begin{footnotesize}
\bibliography{SNLP}
\end{footnotesize}

\end{document}